\documentclass[letterpaper]{article} 

\usepackage[final]{nips_2017}

\usepackage[utf8]{inputenc} 
\usepackage[T1]{fontenc}    
\usepackage{hyperref}       
\usepackage{url}            
\usepackage{booktabs}       
\usepackage{amsfonts}       
\usepackage{nicefrac}       
\usepackage{microtype}      

\usepackage{amsmath,paralist}
\usepackage[ruled,vlined]{algorithm2e}

\newtheorem{theorem}{Theorem}

\newtheorem{lemma}[theorem]{Lemma}

\newtheorem{definition}[theorem]{Definition}
\newtheorem{claim}[theorem]{Claim}

\newcommand{\qed}{\rule{7pt}{7pt}}

\newenvironment{proof-sketch}{\noindent{\bf Sketch of Proof}\hspace*{1em}}{\qed\medskip}
\newenvironment{proof-idea}{\noindent{\bf Proof Idea}\hspace*{1em}}{\qed\medskip}
\newenvironment{proof-of-lemma}[1]{\noindent{\bf Proof of Lemma #1}\hspace*{1em}}{\qed\medskip}
\newenvironment{proof-of-corollary}[1]{\noindent{\bf Proof of Corollary #1}\hspace*{1em}}{\qed\medskip}
\newenvironment{proof-of-corollary-noqed}[1]{\noindent{\bf Proof of Corollary #1}\hspace*{1em}}{}
\newenvironment{proof-of-claim}[1]{\noindent{\bf Proof of Claim #1}\hspace*{1em}}{\qed\medskip}
\newenvironment{proof-of-claim-noqed}[1]{\noindent{\bf Proof of Claim #1}\hspace*{1em}}{}
\newenvironment{proof-attempt}{\noindent{\bf Proof Attempt}\hspace*{1em}}{\qed\medskip}

\newenvironment{remark}{\noindent{\bf Remark}\hspace*{1em}}{\medskip}

\begin{document}
%
\title{Learning Abduction under Partial Observability
	}

\author{ 
 Brendan Juba, \quad
Zongyi Li, \quad
Evan Miller\\ 
Dept. of Computer Science and Engineering\\
Washington University in St. Louis\\
 \text{ \{bjuba, zli, evan.a.miller\} @wustl.edu }
 }

\maketitle



\section{Introduction}
{\em Abduction} is the task of inferring a plausible hypothesis to explain an
observed or hypothetical condition. Although it is most prominently observed in
scientific inquiry as the step of proposing a hypothesis to be investigated, it
is also an everyday mode of inference. Simple tasks such as understanding
stories from \cite{hsam90} and images from \cite{cp86} and \cite{poole90} involve a process of 
abduction to infer an interpretation of the larger events, context, and 
motivations that are only partially depicted. Its significance to AI was first
recognized by \cite{cm85}.

In this work, we consider a {\em PAC-learning}~\citep{valiant84,valiant00} formulation of the combined task 
of {\em learning to abduce}, introduced by \cite{juba16}. In this
formulation, one is given a collection of examples drawn from the prior 
distribution (i.e., example jointly sampled values of attributes) together with 
a condition to explain, represented as a Boolean formula $c$ on the attributes. 
The task is then to propose a formula $h$, which essentially must be a $k$-DNF 
for computational reasons, satisfying the following two criteria:
\begin{compactenum}
\item {\em Plausibility:} the probability that $h$ is satisfied on the prior
distribution must be at least some (given) minimum value $\mu>0$
\item {\em Entailment:} the probability that the condition to explain $c$ is
satisfied, conditioned on the hypothesis $h$ holding, is at least $1-\epsilon$
for some given error tolerance $\epsilon>0$.
\end{compactenum}
By casting the task as operating directly on examples, Juba avoids the problem
of explicitly learning and representing the prior distribution. The main
shortcoming of this formulation is that it assumes access to {\em complete
information,} so any attributes to be invoked in the explanation must be 
recorded in {\em all} of the examples. This is a problem, for example, when we
wish to infer the intentions of characters in stories, which are frequently
either left ambiguous or are assumed to be clear from the given context. It is
also a problem if, for example, we would like to use the abduced hypothesis to
guide further exploration that may include attributes that we previously were
not measuring.

Our work extends Juba's formulation of the abduction task to use partial examples
and draw on declaratively specified background knowledge.
We observe that by using a covering algorithm, it is possible to guarantee
significantly better explanations when a small hypothesis (using relatively few terms) is adequate.
Concretely, when some $r$-term $k$-DNF explanation on $n$ 
attributes has an error rate of $\epsilon^*$, we obtain an error rate of 
$\tilde{O}(r(\log\log n+\log k 
)\epsilon^*)$, in contrast to the bound obtained for the 
state-of-the-art algorithm of \cite{zmj17}, which gave an 
error rate of $\tilde{O}(\sqrt{n^k}\epsilon^*)$ (but does not consider the 
effect of the size of the hypothesis).

\section{Preliminaries}

We work in a standard machine learning model in which the data consists of many
{\em examples,} assigning Boolean values to a variety of {\em attributes.}  For example, if our data is about birds, each bird may correspond to an example and then there can be attributes 
such as: whether the bird has feathers or not, whether it eats bugs or not, and other properties. 

\subsection{Partial Observability}
In the real world, it is hard to require 
each example to contain all of the attributes. So, we want to make inferences with incomplete data.
{\em Partial observability} means that some attributes of examples may be 
unknown. We represent this by allowing the value of each attribute to be 1 
(true), 0 (false), or $*$ (unobserved). For instance, an example $\rho^{(i)}$ 
could be [$x_1=1,\ x_2=* ,\ \cdots ,\ x_n=0$]. (For convenience, we denote 
$\rho^{(i)}$ to be the $i$th example and $\rho_i$ to be the $i$th coordinate of 
an example $\rho$.) 
In our abduction task, we say our partial examples are drawn from such a masked distribution $M(D)$.


\subsection{Implicit Learning}
The main tool to deal with partial observability is {\em implicit learning}. Implicit learning means learning without producing explicit representations.
Given a knowledge base (a set of formulas), and a query
formula, we want to know if the knowledge base can derive the query formula.
The main theorem of implicit learning says, as long as the formulas in a
knowledge base are sufficiently observed in partial examples, we can determine
whether the knowledge base can derive the query {\em without} explicitly
constructing or representing the knowledge base.

\begin{definition}[Witnessed Formula]
Given a partial example $\rho$, we say a 
formula $\phi$ is {\em witnessed} if $\phi|_\rho$ is 0 or 1, where $\phi|_\rho$ means the formula $\phi$ restricted to example $\rho$. 
\end{definition}
Formally, a restricted formula $\phi|_\rho$ is defined recursively: we break down a formula at its logical connectives $(\{\neg, \vee, \wedge \})$ recursively up to single variables, where for the base cases, these singletons are set by the values from the example $\rho$. Restricted formulas' explicit expressions can be computed in linear time.

Informally speaking, we get the restricted formula by plugging in the observed value of the given example, resulting in a (shorter) formula of the unobserved variables. For instance, let $\phi = x_1 \vee x_2$, and in a partial example, $x_1 = 1; x_2 = *$. Then $\phi$ is witnessed (true) even though $x_2$ is not observed. But it could be hard to determine the value for more complex formulas; in general, this may be as hard as deciding if the formula is a tautology, which is NP-hard. Notice that each formula can be either witnessed true, witnessed false, or not witnessed.

\paragraph{Proof system}
Given a knowledge base $KB$ (a set of formulas) and a query formula $\phi$, for our purposes a proof system is an algorithm that can determine whether we can derive $\phi$. 
If we can derive $\phi$, we say $\phi$ is {\em provable} and denote it as $\vdash \phi$. 

A proof system is {\em restriction-closed} if whenever there is a proof of a
formula $\phi$, there is also a proof of $\phi|_\rho$ for any partial assignment
$\rho$. In general, if there is a proof of $\phi|_\rho$ from $KB|_\rho$, we say that
{\em $\phi$ is provable from KB under $\rho$}.
The formal language may be confusing, but the definition is indeed intuitive. Consider the following example:
$\psi_1 = x_1 \wedge x_2$,
$\psi_2 = x_3 \wedge x_4$,
 $\phi = x_1 \wedge x_2 \wedge x_3 \wedge x_4$,
$\{ \psi_1 \wedge \psi_2 \} \vdash \phi$.
 If in $\rho$, $x_1, x_3$ are observed true and $x_2, x_4$ are unobserved, then 
$\psi_1|_\rho = x_2$, $\psi_2|_\rho = x_4$, $\phi|_\rho = x_2 \wedge x_4$,
We thus anticipate, $\{ \psi_1|_\rho \wedge \psi_2|_\rho \} \vdash \phi |_\rho$.

Notice that most common propositional proof systems such as Resolution, (Forward) Chaining, Cutting Planes, and Polynomial Calculus are indeed restriction closed. 

\subsection{DecidePAC Algorithm}
Besides the information we directly witness from the examples, we want to know further what we can infer, given some knowledge base (a set of additional formulas).
From the previous work by \cite{juba13}, we have an algorithm that can tell whether a formula is provable or not.  Given knowledge base $KB$ and partial examples \{$\rho^{(1)}, \cdots, \rho^{(m)} $\} drawn from $M(D)$, for a query formula $\phi$, DecidePAC can tell whether there is a proof of $\phi$ if the knowledge we need is witnessed sufficiently often:
DecidePAC will {\em Accept} if there exists a proof of $\phi$ in from $KB$ and
formulas $\psi_1,\psi_2,\cdots$ that are simultaneously witnessed true with
probability at least $1-\epsilon+\gamma$ on $M(D)$; otherwise, if [KB$\Rightarrow \phi$] is not true with probability at least $(1-\epsilon-\gamma)$, then DecidePAC will $reject$ formula $\phi$.

Notice that there are three different concepts of being true: 1. $observed$ (or $witnessed$), 2. $provable$, and 3. $true$. 
For example, let $t = x_1 \wedge \neg x_2$. In example $\rho^{(1)}$, it is observed that $ x_1 = 1, x_2 =0$, so $t$ is observed to be true in $\rho^{(1)}$; in example $\rho^{(2)}$, $x_1=1$ while $x_2$ is unobserved, but if we assume in KB we have $x_1 \Rightarrow \neg x_2$, then $x_2$ is provable, so $t$ is provable; in example $\rho^{(3)}$, nothing is observed and we know nothing, but in fact, $t$ can be true.
Notice that being observed can imply being provable, and being provable can imply truth. 
We want to bridge from the witnessed values of examples to their ground truth, through logical inference.

DecidePAC was analyzed by Juba using an additive Chernoff bound. We can obtain
an analogous {\em multiplicative} $(1\pm\gamma)$ guarantee by instead using the
multiplicative Chernoff bound:
\begin{lemma}[Multiplicative Chernoff Bound]
Let $X_1,\ldots,X_m$ be independent random variables taking values in $[0,1]$,
such that $E[\frac{1}{m}\sum_iX_i]=p$. Then for $\gamma\in [0,1]$,
\begin{align*}
\Pr\left[\frac{1}{m}\sum_iX_i > (1+\gamma)p\right]&\leq e^{-mp\gamma^2/3}
\qquad \mathrm{and} \qquad
\Pr\left[\frac{1}{m}\sum_iX_i < (1-\gamma)p\right]&\leq e^{-mp\gamma^2/2}
\end{align*}
\end{lemma}

\section{Abduction under Partial Observability}
 Given a query or an event, abduction is the task of finding an explanation for the query or event. An explanation is a combination of some conditions that may
have caused the query. For example, when the query is ``Engine does not run," an explanation can be ``No gas, or key is not turned.''

We require the resulting explanation to satisfy two conditions, {\em ``plausibility''} and {\em ``entailment.''}
Entailment means that when the conditions in the explanation are true, the query should also often be true, or at least rarely false. Thus, the explanation is a (potential) cause of the query. 
Plausibility means the explanation is often true. In other words, for many examples, these conditions are observed. This
suppresses unlikely explanations such as ``A comet hits the car.'' 
which is a valid entailment, but not plausible.

\begin{definition}[Partial Information Abduction]
For any fixed proof system, abduction is the following task: given any query 
formula $c$ and independent partial examples $\{\rho^{(1)}, \ldots, \rho^{(m)}\}$ over a masked distribution $M(D)$,
we want to find a $k$-DNF explanation $h$, such that the explanation $h$ satisfies:
\begin{compactenum}
\item  $\Pr[ \exists t\in h : t\text{ provable under }\rho ]\geq \mu$
{\em (Plausibility) } 
\item $\Pr\left[\neg c \text{ provable} \text{ under }\rho \middle|  \exists t\in h: t\text{ provable}
\text{ under }\rho\right]\leq \epsilon$ {\em (Weak Entailment) }
\end{compactenum}
\end{definition}
Recall, a $k$-DNF explanation $h$ with $r$ terms is in the following form:
$h = t_1 \vee t_2 \vee \cdots \vee t_r,$
where each term 
$t_{i} = \ell_{i_1} \wedge \ell_{i_2} \wedge \cdots \wedge \ell_{i_k}$.
For convenience, we say $t_i \in h$ and $\ell_{i_j}\in t_i$.

\subsection{Choice of Formulation}
We have chosen to relax the condition that $h(x)=1$ in Juba's complete 
information abduction task to the condition that some term of $h$ is provable 
under $\rho$. This is of intermediate strength between $h$ being observed and
$h$ being provable. Provability captures whether or not an agent ``knows'' $t$ 
is true of a given partial example $\rho$. Our choice is somewhat like the
notion of {\em vivid knowledge} by \cite{levesque86}, that the individual literals 
of some definite $t$ should be known. The weaker condition that
merely $h$ is provable is also interesting, but seems much harder to work with;
we leave it as a direction for future work. We could also have relaxed this to
cases where $\neg h$ is not provable, but observe that this includes the cases 
where $h$ is unknown in its favor.  Note that this may ``mix'' many cases where
$h$ was actually false into our estimate of the effect of $h$ occurring, which 
is not desirable, and we anticipate that it would harm the quality of the 
inferences we can draw. 

We made the opposite decision 
for $c(x)=1$, relaxing it to the condition that $\neg c$ is not provable.  The
main reason for this choice is that we wish to not penalize a good $h$ if it
is often impossible to check whether or not $c$ holds. 
We use this liberal notion of entailment for our explanations because the
intended semantics of the task is merely to propose {\em possible} causes given
some tentative partial knowledge of the world, perhaps to guide
further investigation.
At the same time, we
would like to take $\epsilon$ to be very small, so that we can aggressively rule
out $h$'s for which $c$ is frequently known to fail to occur. But, if
we are including the outcome of $c$ being unknown as a ``failure'' of $h$, then
this suggests that in the cases where $c$ is indeed often unknown, then 
$\epsilon$ must be large, even for a good $h$.

\section{Implicit Abduction Algorithm}

A $k$-DNF explanation is actually a disjunction of terms, $h = t_1 \vee t_2 \vee \cdots \vee t_r$. Each term represents a condition, or a possibility. Our goal is to find a formula that covers as many such conditions as possible while
still being a potential cause of the query $c$. 

We observe there is a natural correspondence between our $k$-DNF abduction task and set 
cover: each example of abduction is an element of the set cover problem, and
each term is a set. We say a term covers an example when the term is provable
in that example. The number of examples from the distribution is equivalent to
its frequency or empirical probability with respect to the distribution M(D). If the 
resulting explanation consists of terms that are provable in most of examples, 
then we can conclude that our explanation is provable with high probability. 


\begin{algorithm}
\DontPrintSemicolon
\SetKwInOut{Input}{input}\SetKwInOut{Output}{output}
\Input{Knowledge base KB, , query $c$ and parameters $\mu,\epsilon,\delta,\gamma \in [0,1]$}
\Output{A $k$-DNF explanation $h$}

\Begin{
Initialize $T$ to be the set of all terms of at most $k$-literals.
Draw partial examples $\{ \rho^{(1)},\cdots,\rho^{(m)} \}$ from $M(D)$ for $m = \frac{6}{\mu\gamma^2}\log\frac{2|T|^r}{\delta}\log(\frac{3}{\gamma^2}\log(\frac{2|T|^r}{\delta}))$\\
1. \lForAll{$t\in T$ s.t.
	$\#\{\rho:t\text{ provable under }\rho \wedge \neg c\text{ provable under }\rho\}>\mu\epsilon m$\\}
	{Delete $t$ from $T$.
	}
2. Run {\bf greedy algorithm} for set cover: \\use terms in T to cover a $\mu$-fraction of the examples. Get  $\{t_1, \cdots, t_r\}$.\\
$h \gets t_1 \vee \cdots \vee t_r$\\ 
\Return{$h$.}
}  

\caption{Implicit Abduction}\label{Implicit Abduction}
\end{algorithm}


In the implicit abduction algorithm, 
we enumerate through all possible $k$ literal terms: 
\begin{compactenum}
\item Check all the terms using the same technique underlying DecidePAC:  
We count the number of bad examples where $\neg c$ and $t$ are both provable. 
If the bad examples are more than a $\mu\epsilon$-fraction, then we delete this term.

By the Chernoff bound, all the terms that pass the test then satisfy weak entailment:
the error condition $[\vdash t \text{ and } \vdash \neg c]$ has probability at most $\mu \epsilon(1+\gamma)$. 
\item Then use the greedy algorithm to choose an explanation. If the algorithm can find an explanation covering a $\mu$-fraction of examples, then we can argue explanation has probability larger than $\mu$ by the Chernoff bound.
\end{compactenum}
Thus, if there exists a good explanation, we can find an explanation satisfying entailment and plausibility.

\begin{remark}
If $\mu^*$ is the optimal probability that the terms of a potential explanation 
$h^*$ can be provable, \cite{juba16} showed that a multiplicative 
approximation to $\mu^*$ can be easily found by binary search. We assume that 
such an estimate $\mu$ is given as input.
\end{remark}

\begin{theorem}[Implicit Abduction]\label{main}
Given a query $c$, partial examples $\rho^{(1)},\cdots,\rho^{(m)}$ from a masked distribution $M(D)$, and an efficient restriction-closed proof system with knowledge base $KB$, for constant $k$:

If there exists a $r$-term $k$-DNF ${ h^* = t^*_1 \vee \cdots \vee t^*_r }$ satisfying: 
\begin{compactenum}
\item With probability at least $(1+\gamma)\mu$ over $\rho$ from $M(D)$, 
$\exists t^*_i \in h^*$, such that $t^*_i$ is provable from $KB$ under $\rho$ (Plausibility). 
\item Under $\rho$ drawn from $M(D)$, if some term $t^*$ of $h^*$ is provable, 
then  $\neg c$ is only provable with probability at most $(1-\gamma)\epsilon$. 
(Weak Entailment)
\end{compactenum}

Then, we can find a  $k$-DNF $h$ in polynomial time, such that with
probability $1-\delta$,
\begin{compactenum}
\item $\Pr[\exists t \in h \text{ provable under }\rho]\geq (1-\gamma)\mu$ (Plausibility)
\item $\Pr\left[\neg c\text{ provable}\\\text{ under }\rho
\middle| \exists t \in h\text{ provable }\\\text{under }
\right] < \tilde{O}(r(\log\log n+\log k+\log\log\frac{1}{\delta} +\log \frac{1}{\gamma})(1+\gamma)\epsilon))$ (Weak Entailment).
\end{compactenum}

\end{theorem}



\subsection{Proof of the Main Theorem}

\paragraph{Soundness.} We first show that if the implicit abduction algorithm returns an explanation $h$, then $h$ satisfies weak entailment.
Plausibility will follow from the assumption that a good explanation exists,
so we postpone its discussion to our discussion of completeness, below.

Each term of the explanation is checked by Implicit Learning, so all terms have low error rates: for $\delta'=\frac{\delta}{2{2n\choose \leq k}+4}$,
\begin{claim}\label{claim6}
For our choice of $m\geq\frac{12}{\mu\gamma^2}\log\frac{1}{\delta'}$ we can 
guarantee that with probability $1-\delta/2+2\delta'$, for all terms $t$ that 
pass the first test,
$\Pr[(\vdash t|_\rho ) \wedge (\vdash(\neg c)|_\rho)] < \mu\epsilon(1+\gamma)$
\end{claim}
\begin{proof-of-claim}{\ref{claim6}}
In the Implicit Learning Algorithm, we enumerate through all possible $k$-DNF 
terms over $n$ attributes, so there are at most ${2n \choose \leq k}$ possible 
terms. In the algorithm, for every term $t$ that passes the first test, 
$[(\vdash t|_\rho ) \wedge (\vdash(\neg c)|_\rho)]$ happens in less than a 
$\mu\epsilon$-fraction of the examples. By the multiplicative Chernoff bound, 
when we take enough examples, we will be able to guarantee that $\Pr[\#\{\rho:
(\vdash t|_\rho) \wedge (\vdash(\neg c)|_\rho)\} < (1-\gamma/2)(1+\gamma)\mu\epsilon ]<
\delta'$,  
i.e., any term with at most $\mu\epsilon$ bad examples has error at most
$(1+\gamma)\mu\epsilon$ with high probability.
For each term, the Chernoff bound requires $\frac{12}{\mu\gamma^2} \log(\frac{1}{\delta'})$ examples to be correct with probability $1-\delta'$. We have chosen
$\delta'$ so that after a union bound over the terms we get $\delta/2-2\delta'=
{2n\choose\leq k}\delta'$. Thus, $m \geq \frac{12}{\mu\gamma^2}\log\frac{1}{\delta'}$ examples suffice.
\end{proof-of-claim}


\paragraph{Completeness.}
We just proved that every output satisfies weak entailment with
probability $1-\delta/2+2\delta'$. Now, we want to show if there is an optimal r-term k-DNF explanation  ${ h^* }$ satisfying 
\begin{compactenum}
\item (Plausibility) for a $(1+\gamma)\mu$-fraction of examples, some term 
$t\in h^*$ is provable, and
\item (Weak Entailment) if some $t\in h^*$ is provable, then with high probability $\neg c$ is not provable
\end{compactenum}
\noindent
then we are able to find a good solution that satisfies plausibility and weak entailment.

\begin{claim}\label{claim8}
If there exists a solution $h^* = t^*_1 \vee t^*_2 \vee \cdots \vee t^*_{r}$ such that 
[$\neg c$ is provable when some $t^*_i$ is provable]
has probability at most
$(1-\gamma)\mu\epsilon$, 	 
then all these terms $t^*$ can pass the first test with probability
$1-\delta'$.
\end{claim}

\begin{proof-of-claim}{\ref{claim8}}
We are given that $\Pr[\ [(\vdash t^*_1|_\rho) \wedge (\vdash(\neg c)|_\rho)] \vee
\cdots\vee [(\vdash t^*_{r}|_\rho) \wedge (\vdash(\neg c)|_\rho)]\ ] < 
(1-\gamma)\mu\epsilon$.
By a Chernoff bound, for our choice of $m$, $[(\vdash t^*|_\rho) \wedge 
(\vdash(\neg c)|_\rho)]$ happens for any
$t^*$ in $h^*$ in less than $\mu\epsilon$-fraction of examples with probability,
$1-\delta'$ so all these terms $t^*$ pass the first test.
\end{proof-of-claim}

Next, we show the number of terms $r'$ is controlled, since $r'$ depends upon the solution of the set cover problem.
\begin{claim}\label{claim4}
If there exists a solution $h^* = t^*_1 \vee t^*_2 \vee \cdots \vee t^*_{r}$ that satisfies 
\begin{compactitem}
\item $\Pr[\exists t\in h^*:\ \vdash t|_\rho]\geq (1+\gamma)\mu$ 
\item $\Pr[\ \vdash(\neg c)|_\rho\ |\ \exists t\in h^*:\ \vdash t|_\rho\ ] < 
(1-\gamma)\epsilon$
\end{compactitem}
then Implicit Abduction finds an $h$ using at most $r' = r \log(\mu m)$ terms 
such that $\#\{\rho\ :\ \exists t\in h , \vdash t|_{\rho}\}>\mu m$. 
Furthermore, by a union bound on the error of each term, $\Pr[(\exists t\in h:\ 
\vdash t|_\rho ) \wedge (\vdash(\neg c)|_\rho)] < \mu\epsilon(1+\gamma)$.
We thus find that with probability at least $1- \delta $ 
$h$ satisfies plausibility with $(1-\gamma)\mu$ and weak entailment.
\end{claim}
\begin{proof-of-claim}{\ref{claim4}}
Following Claims \ref{claim6} and {\ref{claim8}}, with probability at least 
$1-\delta/2+\delta'$, all terms $t^*$ in $h^*$ can pass the first 
test, so they are available for set cover. Moreover, by another Chernoff bound,
since at least one of the terms of $h^*$ is provable with probability 
$(1+\gamma)\mu$ in each example, with probability $1-\delta'$ at least one of 
the terms is provable in at least a $\mu$-fraction of the $m$ examples. Thus, 
there is a set of $r$ terms (the terms of $t^*$) that 
pass these tests and indeed cover a $\mu m$ examples.
For the greedy algorithm, if Opt ($h^*$) covers $\mu m$ examples using $r$ 
sets, then our greedy algorithm can find a cover using $r' = r \log(\mu m)$ sets that also 
covers $\mu m$ examples~\cite{slavik97}.

Recall that $h = t_1 \vee \cdots \vee t_{r'}$. For each term, by Claim~\ref{claim6}, 
$\Pr[(\vdash t|_\rho ) \wedge (\vdash(\neg c)|_\rho)] < \mu\epsilon(1+\gamma)$, 
so if take an union bound over the terms of $h$, the error, 
$\Pr[\exists t\in h (\vdash t|_\rho) \wedge (\vdash(\neg c)|_\rho)]$, is at most 
$r'\mu\epsilon(1+\gamma)$ in total.
If we plug in $r' = r \log(\mu m)$, the resulting error is $O(r\log(\mu m)(1+\gamma)\mu\epsilon)$.

To see that the returned $h$ satisfies plausibility, we consider a Chernoff bound for the fraction of examples in which each possible $r'$-term $k$-DNF has
a provable term with $\hat{\delta} = \delta/2|T|^{r'}$. So when we take a union bound on all $k$-DNF explanations, any $r'$-term explanation found will actually have plausibility $(1-\gamma)\mu$ with probability $1-\delta/2$. Therefore, it suffices to have
\[m \geq  \frac{3}{\mu\gamma^2}\log\frac{2|T|^{r'}}{\delta}
\quad\mathrm{or}\quad 
m\geq\frac{3r\log(\mu m)}{\mu\gamma^2}\log\frac{2|T|}{\delta}.\]
Here we apply the inequality
\begin{lemma}
For $a\geq 1$, if $x\geq 2a\log a$, 
then $x\geq a\log x$.
\end{lemma}
By plugging in $x=\mu m$ and
$a= \frac{3r}{\gamma^2}\log\frac{2|T|}{\delta}$, 
we get 
   $m\geq \frac{6 r}{\gamma^2 \mu}\log(\frac{2|T|}{\delta})\log(a)$
examples suffice.
Here, $\log a$ is dominated by other terms, so we get $m=\tilde{O}(\frac{r}{\gamma^2\mu}\log\frac{n^k}{\delta})$.

Since we condition on some $t\in h$ provable and 
$\Pr[\exists t\in h\text{ provable under }\rho]>(1-\gamma)\mu$,
\begin{align*}
\Pr & [\ \vdash(\neg c)|_\rho\ |\ \exists t\in h:\ \vdash t|_\rho\ ]\\
&= \Pr[(\exists t\in h \vdash t|_\rho ) \wedge (\vdash(\neg c)|_\rho)]/ \Pr[\exists t\in h:\ \vdash t|_\rho]\\
&<O(r\log(\mu m)(1+\gamma)\mu\epsilon/\mu)\\
&= O(r\log(\mu m)(1+\gamma)\epsilon)
\end{align*}
and thus, we indeed find an $h$ satisfying weak entailment with the claimed
error rate with probability $1-\delta$.
\end{proof-of-claim}

Finally, when we plug in  $m = \tilde{O}(\frac{r'}{\mu\gamma^2}\log\frac{3(2n)^k}{\delta})$,
\begin{align*}
&O(r\log(\mu m)(1+\gamma)\epsilon)= \tilde{O}(r\log(\frac{\mu r}{\mu \gamma^2}\log\frac{n^k} {\delta} )(1+\gamma)\epsilon) \\
&\ = \tilde{O}(r(\log\log n+\log k+\log\log\frac{1}{\delta} +\log \frac{1}{\gamma})(1+\gamma)\epsilon)
\end{align*}
We conclude that $\Pr[\ \vdash (\neg c)|_\rho\ |\ \exists t\in h:\ \vdash t|_\rho\ ] < \tilde{O}(r(\log\log n+\log k+\log\log\frac{1}{\delta} +\log \frac{1}{\gamma})(1+\gamma)\epsilon))$ with probability $1-\delta$.

\paragraph{Running time.}
The  test is run for each term of size at most $k$, of 
which there may be $\sim n^k$. And DecidePAC runs in time polynomial in
$n$, $|\varphi|$, $|KB|$ (the running time of the underlying algorithm for the
proof system), and $\frac{1}{\gamma^2}\log\frac{1}{\delta}$; the overall
running time is also polynomial, as needed.\hfill\qed

\section*{Acknowledgements}
B.~Juba and E.~Miller were partially supported by an AFOSR Young Investigator
Award.

\bibliography{robust}
\bibliographystyle{aaai}
\end{document}